\documentclass[AMA,STIX1COL]{WileyNJD-v2}

\articletype{Article Type}%

\received{26 April 2016}
\revised{6 June 2016}
\accepted{6 June 2016}

\usepackage{paralist}
\newcommand{\xhdr}[1]{\vspace{2pt}\noindent\textbf{#1} }

\begin{document}

\title{Improving Users' Mental Model with Attention-directed Counterfactual Edits}

\author[1]{Kamran Alipour*}

\author[2]{Arijit Ray}

\author[2]{Xiao Lin}

\author[2]{Michael Cogswell}

\author[1]{Jurgen P. Schulze}

\author[2]{Yi Yao}

\author[2]{Giedrius T. Burachas}

\authormark{ALIPOUR \textsc{et al}}

\address[1]{\orgname{University of California, San Diego}, \orgaddress{\city{La Jolla}, \state{California}, \country{USA}}}

\address[2]{\orgname{SRI International}, \orgaddress{\city{Princeton}, \state{New Jersey}, \country{USA}}}

\corres{*Kamran Alipour, 9500 Gilman Dr, La Jolla, CA 92093. \email{kalipour@eng.ucsd.edu}}

%\presentaddress{9500 Gilman Dr, La Jolla, CA 92093}

\abstract[Summary]{In the domain of Visual Question Answering (VQA), studies have shown improvement in users’ mental model of the VQA system when they are exposed to examples of how these systems answer certain Image-Question (IQ) pairs \cite{alipour2020study}. In this work, we show that showing controlled counterfactual image-question examples are more effective at improving the mental model of users as compared to simply showing random examples. We compare a generative approach and a retrieval-based approach to show counterfactual examples. We use recent advances in generative adversarial networks (GANs) to generate counterfactual images by deleting and inpainting certain regions of interest in the image. We then expose users to changes in the VQA system’s answer on those altered images. To select the region of interest for inpainting, we experiment with using both human-annotated attention maps and a fully automatic method that uses the VQA system’s attention values. Finally, we test the user’s mental model by asking them to predict the model's performance on a test counterfactual image. We note an overall improvement in users’ accuracy to predict answer change when shown counterfactual explanations. While realistic retrieved counterfactuals obviously are the most effective at improving the mental model, we show that a generative approach can also be equally effective.}


\maketitle

\section{Introduction}
\label{sec:intro}
% Growing use of AI
%The latest advancements in the field of AI are leading us to a new era of smart automation where machines take over human tasks with an unprecedented speed. AI solutions are now extended to a wide range of applications including medicine, defense, and education. Among the latest developments in this field, deep learning has provided us with AI machines with high accuracy. Throughout recent years, the development of deep neural networks (DNNs) based on a vastly growing population of datasets resulted in a surge in the use of such models and also exposed their incompetence due to either model or data deficiencies. On the other hand, the black-box nature of these machines often becomes a source of confusion, especially in failure modes. This issue leads to a sense of uncertainty on whether the pace of AI expansion is appropriate and proportional to its competency and capabilities.

%why is it important
With the growing application of AI in high-risk domains, it is important for human users to understand the extent and limits of AI system competencies to ensure efficient and safe deployment of such systems.
While deep neural networks have made impressive strides, they are notorious for being unpredictable to a human user as to when they succeed or fail in producing correct outputs.
Hence, we need effective approaches to improve the end users' mental model of the deep neural network-based AI systems.  
%While over-trusting an incompetent AI machine can be catastrophic in sensitive situations, lack of trust renders these machines inefficient. 

%how can one solve this
%Hence, human trust in AI should be calibrated based on a realistic view of AI's capabilities in different aspects of the task.
There has been work in literature \cite{chandrasekaran2018explanations} that shows humans can improve their mental models by mere exposure to the system predictions for a variety of inputs. A mental model is a person's internal representation of the AI system she is interacting with and ideally builds a correct understanding of the way that system works \cite{rutjes2019considerations}.
%what do we do
In this paper, we explore the various ways we can present such explanatory additional input-output examples to a user to maximize their mental model improvement. We ask the question: are certain examples of how the machine behaves better than other examples to teach humans when to trust the model and when not to? 
We examine the effect of exposing the users to explanatory examples where the inputs are changed in a controlled manner in order to better observe how the machine output changes to controlled changes in the input. We call these controlled changes in input, ``counterfactuals''. We hypothesize that such controlled changes in the examples shown are better for mental model improvement than showing random examples.

\begin{figure}[t]
\centering
\includegraphics[width=\columnwidth]{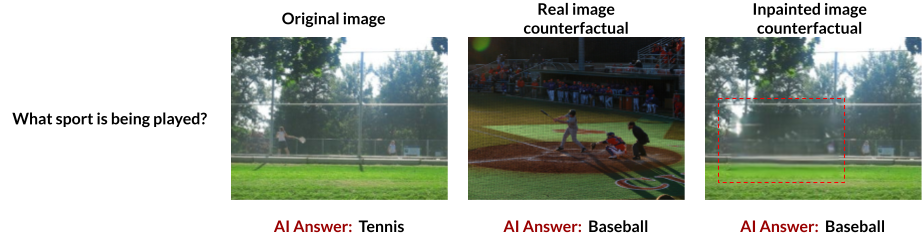}
\caption{While alternative real images may present a convincing counterfactual case for a VQA model, they are expensive to harvest and also often incapable of selecting specific features. In this sample, while the real-image counterfactual may suggest that the AI agent is correctly capturing the type of sport, the in-painted counterfactual suggests that the change in the answer is not necessarily correlated to the changes in the input.}
\label{fig:intro_figure}
\end{figure}

Many approaches \cite{alipour2020impact, ray2019lucid, chandrasekaran2018explanations} to improving mental models also focus on using explanations that aid the user in understanding how a deep network arrived at a certain conclusion. 
While many existing explanation approaches such as attention maps attempt to provide insights into the inner working of AI machines, they don't necessarily convey the causal chain of inference that happens in the algorithm \cite{ray2021knowing}.
As a result, the research community actively seeks novel explanation modalities to probe the causality of AI as this form of explanation can resonate better with human logic. Humans tend to learn better from explanations that easily convey when a machine is about to be correct and when not \cite{ray2021knowing}. Among different techniques, showing counterfactual examples are considered \emph{human-friendly} explanations because they are contrastive and also selective when showing the feature changes \cite{molnar2020interpretable}. Counterfactuals provide the opportunity for the user to explore the range of responses from AI as they manipulate certain features of the inputs and the conditions.

% AR: Maybe move this to related work? Seems more apt there. 
% The philosophical theories of causation and counterfactuals
%In recent decades, the contrastive nature of counterfactuals has drawn the attention of researchers in the field of philosophy into this form of explanation \cite{kment2006counterfactuals,ruben2004london}. Woodward introduces a more general rendition of this idea in his interventionist theory of causation as he argues that \emph{X} is a cause of \emph{Y} if there is a possible intervention on \emph{X} that changes \emph{Y} \cite{woodward2005making}.

% Our work
In this paper, we focus on improving users' mental models by generating counterfactual explanations for the task of visual question answering (VQA) - answering natural language questions asked about images. Specifically, we compare various methods of generating counterfactual examples to maximize a user's accuracy in predicting when a model is about to fail or succeed.  
%Within this task the AI algorithm receives a question and image and attempts to answer that question accordingly. VQA machines combine the features in two widely different domains: language and vision. The complexity of feature space in these models make them extra hard to interpret and explain. Our work is focused on producing counterfactual explanations for the vision aspect of the VQA model.\\
In this setting, given an image-question pair, a counterfactual explanation is showing the output of the model for the same question but on a different image where the answer should be different. For example, as shown in Figure \ref{fig:intro_figure}, for the question ``what sport is being played?" on the original image of playing tennis, the counterfactual examples could be showing the answer of the model on an image where someone is playing baseball (middle image), or where a tennis racket is absent (rightmost image).   
%In creating effective counterfactuals, choosing the right collection of features to alter is a key step. The quality and helpfulness of a counterfactual explanation are heavily dependent on the way the features are selected, altered, and fed to the model. Unlike many previous approaches that use AI outputs, we propose a mechanism based on human attention to target and alter visual features. We believe a counterfactual explanation produced with this method can produce human-friendly insights on the way AI perceives visual input.\\

Specifically, we compare a retrieval-based approach and a GAN-based approach to generate counterfactual images for a given question. For example, as shown in Figure \ref{fig:intro_figure}, we can generate a counterfactual image (an image where the answer may be different from the original image) by either retrieving an image where the answer is different (middle image) or by removing the tennis racket using a GAN network (rightmost image).  
Our automated approach using a GAN provides the opportunity to produce counterfactuals at scale and evaluate their effectiveness on a large population of AMT workers. 

One major challenge of automatically generating counterfactual images is that we are limited by the capability of current GAN models. We chose to use an in-filling network \cite{chang2018explaining} to remove parts of the image since we observed that current networks can achieve this with a reasonable performance. 
Limited by the capability of only being able to remove parts of images, we need to decide the most effective parts of images to remove to generate counterfactual examples that help users to learn the idiosyncrasies of the model in order to improve their mental model.
To this regard, we experiment with using attention maps (heatmaps that point to where a machine looks at while answering the question) to decide relevant and irrelevant parts of the image to remove to generate counterfactual images. 

%Our user studies are intricately designed to test user's ability in predicting AI's behavior. We use these prediction accuracy metrics to evaluate the role of explanations on user's mental model of the AI system and quantify the helpfulness our counterfactuals.
%Finally, we show the counterfactual images to users and measure their accuracy in predicting when a model is about to fail or succeed to measure their mental model.  

\noindent In summary, our contributions include:\\
\xhdr{1) proposing effective ways to generate counterfactual examples:} We outline several ways of generating counterfactual images. Specifically, we compare a retrieval-based method and an automated GAN-based method to generate counterfactual images.\\
\xhdr{2) we evaluate empirically the effectiveness of counterfactual examples relative to providing random examples.}  We show an improvement in the mental model of users when showing controlled counterfactual examples as compared to simply showing random examples or no examples at all.

In the following sections, we first take a look at the related work. We then discuss the methodology behind our approach. We then cover the details for our hypotheses and experimental designs. Finally, we provide the results from our studies and discuss our interpretations. 
\section{Related work}
\label{sec:relatedwork}

\xhdr{VQA/Explanations} Our approach is based on interactions with a visual question answering (VQA) \cite{antol2015vqa} machine. The use of attention-based layers and explanations in VQA has been a highly popular approach \cite{teney2018tips,10.1007/978-3-319-46478-7_28,fukui2016multimodal,alipour2020impact}.
Previous work in the attention-based VQA includes attempts to improve the attention mechanism through co-attention between image and question \cite{DBLP:journals/corr/LuYBP16}, or a combined bottom-up and top-down to compute object-level attentions \cite{anderson2018bottom}. In recent work, Peng \emph{et al.} \cite{peng2020mra} propose a Multi-modal Relation Attention Network (MRA-Net) model with textual and visual relation attention for higher performance and interpretability. Patro \emph{et al.} utilize adversarial training of the attention regions as a two-player game between attention and explanation\cite{patro2020explanation}.  We adopted a VQA model similar to what was proposed by Alipour \emph{et al.} \cite{alipour2020impact} where the attention is derived from a transformer model \cite{devlin2018bert}.

\xhdr{Counterfactuals} Counterfactual examples have also been used to explain image classifiers \cite{chang2018explaining, pmlr-v97-goyal19a}. They have also been used in an optimization process where \cite{wachter2017counterfactual} proposed a loss function to find the minimum changes in the input that results in a change in the output of a classifier.
Using counterfactual images as explanations can also be thought of as the visual equivalent to observing VQA behavior by rephrasing the question and checking if the model responds consistently \cite{ray2019sunny, selvaraju2020squinting, agarwal2020towards}. Hence, such counterfactual images hint at how consistent these models are to users, and that aids in their mental model improvement. 

% Explanation evaluation
%Regardless of the methodology, the explanations for an AI system should be put to the test in practice. Evaluating the effectiveness of explanations has been a topic of discussion among the XAI community in recent years. The prior work tackles this challenge through user studies to assess the effectiveness of the XAI systems in building a better mental model for the users.

\xhdr{Mental model evaluation} Some of the previous studies introduce metrics to measure trust with users \cite{cosley2003seeing,ribeiro2016should}, or the role of explanations to achieve a goal \cite{kulesza2012tell,narayanan2018humans,ray2019lucid}. Dodge et al. investigated the fairness aspect of explanations through empirical studies \cite{10.1145/3301275.3302310}. Lai and Tan \cite{lai2019human} examined the role of explanations in user success within a spectrum from human agency to full machine agency. Lage et al. proposed a method to evaluate and optimize human-interpretability of explanations based on measures such as size and repeated terms in explanations \cite{lage2019evaluation}. Other approaches measured the effectiveness of explanations in improving the predictability of a VQA model \cite{chandrasekaran2018explanations,alipour2020study}.\\
In this work, we develop a series of user studies with a subject population of lay users with minimum knowledge about AI. The experiments are designed to investigate effective methods to produce counterfactuals that can improve the user's mental model of a VQA system.
\section{Method}
\label{sec:method}
In this section, we first describe our VQA model and then explain how we generate counterfactual images using a GAN. 

%In this section we cover the details of our approach in evaluating the explanations on user's ability to predict the AI's behaviour.
%We first provide some details about the AI agent used in the experiments. The model details are followed by the description of algorithms that generate the counterfactual explanations.
\subsection{VQA Model}
\label{sec:method_model}
\noindent Our VQA model is trained based on the VQA 2.0 dataset \cite{goyal2017making} and is capable of answering questions about images in textual format. The model is a transformer-based neural network that can parse a combination of visual and textual embeddings from an image and question.  The model encodes the image into a $49\times512$ feature map with the help of ResNet152\cite{he2016deep}. The objects in the image are also encoded separately into a $36 \times 512$ feature map using a Region Proposal Network \cite{he2017mask}. The model accepts questions with a maximum length of 30 words and all questions below this limit are padded with 0's. The question array is also embedded into a $30 \times 512$ vector of features.

The model employs transformer-based attention layers that receive all the visual, object, and textual features in the concatenated shape of $115$ ($30+36+49$) tokens. The transformer is comprised of four layers with 12 heads in each layer. Consequently, the model can provide an attention tensor between these tokens with a $4\times 12\times 115\times 115$ dimension.
The model provides its prediction as a softmax probability distribution over 3129 answer choices from the attention-weighted feature values.
%Similar to the previous approaches \cite{alipour2020impact}, 
%To visualize an attention map, we compute the overall attention by averaging the weights on the visual features by all input tokens in all heads of the last layer of the transformer.\\

For our experiments, we use a subset of the VQA 2.0 validation dataset. 
We first show the VQA model's answer to the original images and questions from this subset. 
For each example, we also show the answer to two counterfactual images for the question to the user. 
We finally test the user's mental model by asking the user to predict the correctness of the answer on a test image for the same question. 
We will now describe how we generate counterfactual images. 

%We also collect AI's answer to the counterfactual forms of these inputs where we modified the images by in-painting parts of the image.\\

\subsection{Generating Counterfactual Images}

%In our studies, we specifically investigate the role of objects in the image and their impact on VQA system's prediction in a counterfactual scenario. 
We generate counterfactual images to serve as examples of VQA behavior under differing inputs to improve the mental model of users. 
For example, a user who sees a VQA not counting oranges properly when changing the number of oranges in a picture and asking ``How many oranges?'' will learn that the VQA model has a low accuracy for counting oranges. 
This sort of mental model improvement might not have been noticed if we presented the user with only one counting example and other random examples of images and questions. 
In our study, we focus on altering objects in the image for a certain question. 
Specifically, we use a GAN which has been trained to in-paint areas of the image such that it looks natural\cite{chang2018explaining}.
When asked to in-paint an area of a certain object in an image, such a GAN would usually omit the object and in-paint its content that matches the background/surrounding scenery. 
We use such an approach to remove objects from the scene. 
However, such approaches are currently noisy and we often note artifacts in the image that make it seem unnatural.
Hence, we limit the size of all in-painting bounding boxes to 10\% to 20\% of the whole image area.
%We seek an algorithm that can identify certain areas of the image and then in-paint them to produce a counterfactual input for the VQA model.
%For the in-paint process, we use a GAN that can in-paint a certain bounding box within image and fill it with content assimilated to the background. filling a bounding box in a real image with generative content is not a trivial task and the GAN model can introduce artifacts to the results. To avoid this artifacts which can create unwanted impact on the VQA model's answer, we limit the size of all in-painting bounding boxes to 10\% to 20\% of the whole image area.\\

\xhdr{How to choose objects to in-paint}

\noindent The ultimate goal for this algorithm is to generate counterfactual explanations that are helpful to the users in predicting AI's response. 
Given the diverse combinations and interactions between objects in a real scene, it is not obvious how to define an algorithm to select and alter the objects from images to maximize mental model improvement.
In our approach, we use attention maps - heatmaps that convey the important regions of the image for answering the question - to decide objects to remove in the image. 
We use two different sources of attention to identify the in-paint candidates and then produce the counterfactual images based on them:
\begin{compactenum}[--]
\item Human annotated attentions for the image-question pair, which come from the human attention dataset \cite{das2017human}. 
\item The attention layers from the AI system. As described in Section \ref{sec:method_model}, our VQA system has multiple layers of attention that weigh the image and question features. We select the weights from the last layer (averaging over the transformer heads) to display the attention values over the image regions. The attention values over the image regions are also computed as the average weight over all question tokens.  
\end{compactenum}

Based on the above-computed attention maps, we generate two counterfactual images- 1) we remove a box that falls in a region of high attention, and 2) we remove a box that falls in a region of low attention. This ensures we remove a relevant and irrelevant object in the image to introspect how the VQA model's answer changes. Based on this observation, a user can hypothetically learn whether the VQA model is behaving rationally or not. To select the low and high attention boxes, we employ a threshold that first segments the high and low attention regions from the attention map. The bounding boxes surrounding these regions provide the in-paint area. In cases where the bounding boxes are outside the limits (10\% - 20\% of the image area), the proposed box is scaled to a size within the range.

\begin{figure*}[h]
\centerline{
\includegraphics[width=400pt]{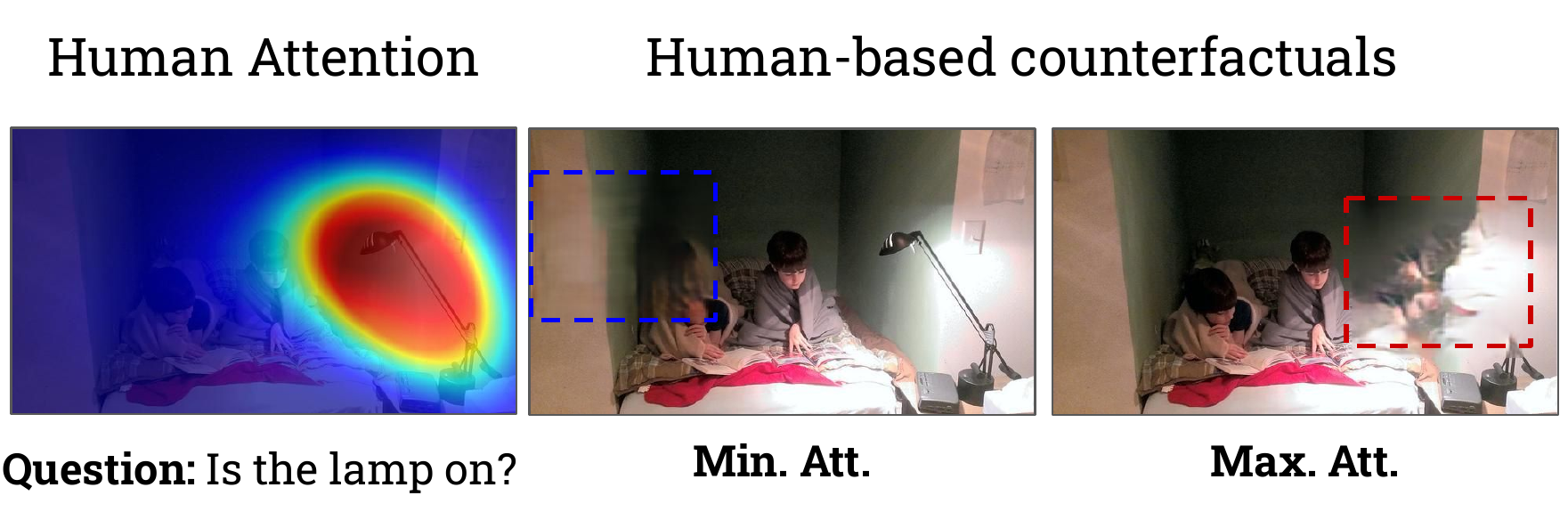}}
\caption{Generating counterfactual images based on human annotation attentions. The algorithm first identifies the most attended and least attended bounding boxes in the image and then applies the GAN to in-paint those bounding boxes and produce the counterfactual images.}
\label{fig:hat_cf_generation_procedure}
\end{figure*}
\section{Experimental Settings}
We conduct experiments to quantify the improvement in the mental model for users after being exposed to counterfactual explanations. 
We measure the user mental model by asking them to predict the answer-change or the correctness of the answer for a given image-question (IQ) pair, similar to concurrent studies on user mental model evaluation \cite{chandrasekaran2018explanations, alipour2020study}. 
We use the Amazon Mechanical Turk platform to recruit users for our study. We recruit workers located in the United States (due to IRB regulations) and who exceed 98\% approval rating on over at least 50 such human-intelligence tasks (HITs).  
% Experiment design, framework, objectives
%As mentioned earlier, our experiments are designed to quantify the impact of counterfactual explanations on user's understanding of AI operation. 
%We conducted multiple user-studies under the Amazon Mechanical Turk (AMT) platform to evaluate the helpfulness of different counterfactual explanations on users prediction accuracy.\\

\noindent In our study, each user goes through 1 HIT which consists of 20 episodes of IQ pairs. 
In each episode, the users first see the VQA model's response to the original Image-Question (IQ) pair.
Based on their group configuration, then they may or may not see two counterfactual forms of the original image and AI's response to the original question for those counterfactual images. 
In the evaluation section of each episode, users attempt to predict AI's response to a test image for the same question. We quantify user's mental model states based on their accuracy in predicting AI's response.

%\noindent We study the helpfulness of counterfactual explanations in two major steps. In the first step, we study the impact of showing counterfactual examples compared to showing random examples or no examples. We then evaluate the kind of counterfactual examples that maximize mental model improvement. Specifically, we experiment with attention map directed generated counterfactual images and retrieved counterfactual images.

We use two tasks to measure a user's mental model - a) answer-change prediction to see if users can predict if the answer will change when a certain object is removed, and b) answer correctness prediction on a real test image based on the lessons learned from counterfactual examples. 
\vspace{-2mm}
\subsection{Answer-change prediction}
In this setting, we show an IQ pair to a user, the VQA model's answer on the original IQ. 
One group of users - Counterfactual Group (CF Group) - sees two examples of objects being removed from the image and the VQA model's answer on these two altered images. 
The Control Group of users see no such altered examples. 
Finally, both the groups of users are presented with another new object removed from the same image and are asked to predict if the VQA model's answer for that image will change from the original image or not. 
%First, we probe the role of counterfactuals by examining users' prediction abilities in the presence and absence of counterfactual samples. In both scenarios users first see AI's answer to an IQ pair and they attempt to predict whether the AI's answer would change if the image is in-painted in a certain area while the question stays the same. We consider the user's prediction ability in this task as an indicator of the user mental model. In the baseline groups of the study, users only see the original IQ and AI's answer to that. They then attempt to predict whether AI's answer will change if a part of the image is in-painted. On the other hand, in the explanation groups, users also see two samples of the in-painted image and AI's answer to the question for those in-painted images. (figure \ref{fig:exp_config_noexp_vs_in-paint_anschange}).\\
\begin{figure*}[h]
\centerline{
\includegraphics[width=500pt]{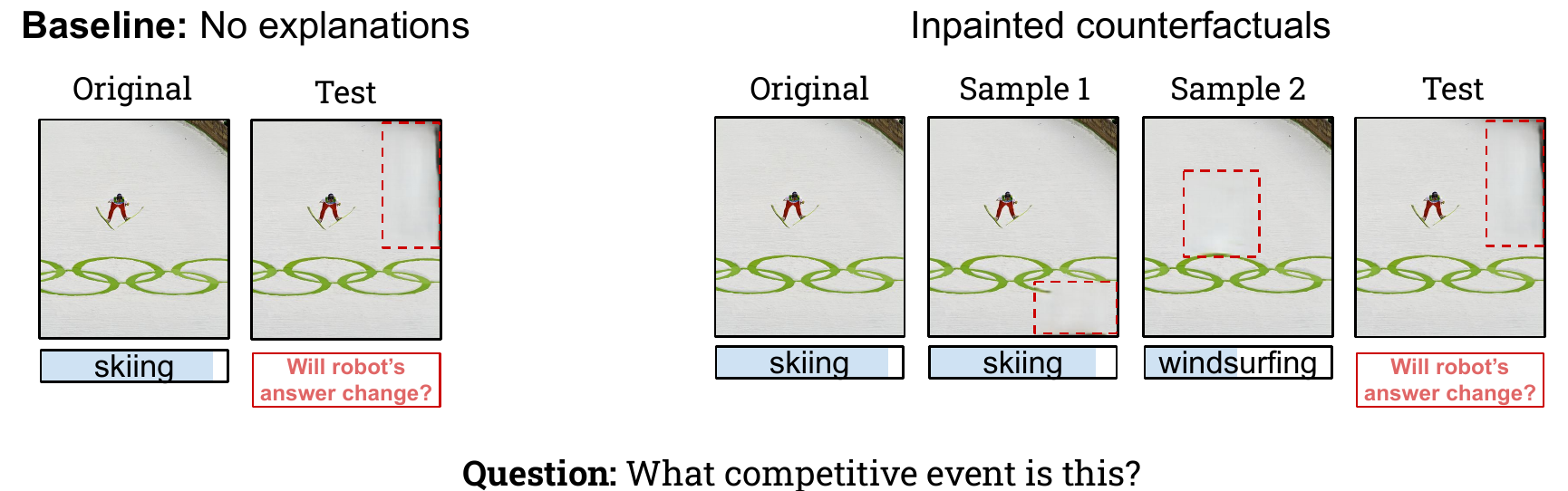}}
\caption{The interfaces for the experiments that evaluate the impact of in-painted counterfactuals for the task of answer-change prediction. Users in both groups are evaluated based on the same in-painting patterns. While the users in the counterfactual groups can utilize the counterfactual samples in their prediction, the baseline group attempts to predict the answer-change merely based on the original IQ response. For the input and sample images, users see AI's top answer along with its probability (blue bar beneath the answers).}
\label{fig:exp_config_noexp_vs_in-paint_anschange}
\end{figure*}

In the experiment, the counterfactuals were generated based on human attention \cite{das2017human} annotations from the VQA-HAT dataset. 
%The attention annotations in this dataset are collected from multiple users who were asked to highlights those areas of the image they look at while answering the question. 
%We defined human attention in our application as the average of three attention annotation maps provided for each IQ in VQA-HAT dataset.\\ 
Note that the object removed in the test image is always different from the objects removed in the counterfactual examples shown. 
%In each pair of IQ, users in the counterfactual group were provided with two samples of counterfactuals where some area of the original image was in-painted. 
We do this by choosing separate regions of minimum (min), maximum (max), or medium (mid) attention based on the human-attention values on the image. 
While the Min and Max regions are determined by extracting the areas from the two sides of the attention spectrum (see figure \ref{fig:hat_cf_generation_procedure}), Mid area is identified by avoiding the overlap with Min and Max and also maintaining the minimum attention possible. This procedure of in-painting assures the minimum overlap among the sample and the test in-paintings and therefore minimizes the chance of overlap between counterfactual samples and test images. 
While testing the users, we randomly choose to show two of min, mid, and max as counterfactual examples and test on the unseen third. The group \textbf{CF-MinAtt} shows mid and max attention as samples and tests on the image with the min attention region removed. Similarly, \textbf{CF-MidAtt} tests on the image with the mid attention region removed, and \textbf{CF-MaxAtt} tests on the image with the max attention region removed. 
%This experiment examines the impact of sample counterfactuals on user performance as compared to a scenario where the user does not have that information before prediction.

%\begin{center}
%begin{table*}[h]%
%\caption{Counterfactual groups for answer change prediction. Each of these group's performances are compared to their counterparts where there are no sample images provided.\label{table:anschangeprediction_groups}}
%\centering
%\begin{tabular*}{350pt}{@{\extracolsep\fill}lccD{.}{.}{3}c@{\extracolsep\fill}}
%\toprule
%\textbf{Group} & \textbf{Sample images in-painting}  & \textbf{Test image in-painting} \\
%\midrule
%CF-MinAtt & Mid \& max attention areas & Min attention areas \\ 
%CF-MidAtt & Min \& max attention areas & Mid attention areas \\ 
%CF-MaxAtt & Min \& mid attention areas & Max attention areas \\ 
%\bottomrule
%\end{tabular*}
%\end{table*}
%\end{center}
\vspace{-2mm}
\subsection{Answer correctness prediction}

In the second set of experiments, we provide a more realistic setting to evaluate the user mental model. Instead of predicting an answer-change for a counterfactual test image, the users now attempt to predict whether the model will answer the same question correctly for a different test image. Since the test images are also selected from the IQ pairs in the VQA dataset, that guarantees that the test image is relevant for the question asked.\\
We define four groups (shown in table \ref{table:accuracyprediction_groups}) to check whether counterfactual examples improve users' mental models to be able to predict the model's correctness on an unseen test image: 
\begin{compactenum}[--]
\item Control Group (\textbf{CG-NoExp}) sees no explanations and just the VQA model's answer on an IQ pair. 
\item the counterfactual group is either based on counterfactual images generated using human-annotated attention (\textbf{CF-HAT}) or VQA model's attention (\textbf{CF-AIAtt}). These groups are to examine how the process of generating counterfactual images affects the mental model.  
\item a group that sees retrieved real counterfactual images (\textbf{CF-AltImg}). We retrieve images that are relevant to the question but have a different answer from the VQA dataset. We can think of these as ideal counterfactual examples. The performance of this group compared to the CF-HAT and CF-AIAtt groups would tell us if generated counterfactuals (CF-HAT and CF-AIAtt) can be used in place of real counterfactual data to reduce dataset collection costs. 
\item a group that sees random IQ pairs instead of counterfactuals (\textbf{CG-RandIQ}). This group is to understand how much we gain from simply presenting two samples of random IQ pairs instead of two counterfactual IQ examples.
\end{compactenum}

\begin{figure*}[h]
\centerline{
\includegraphics[width=500pt]{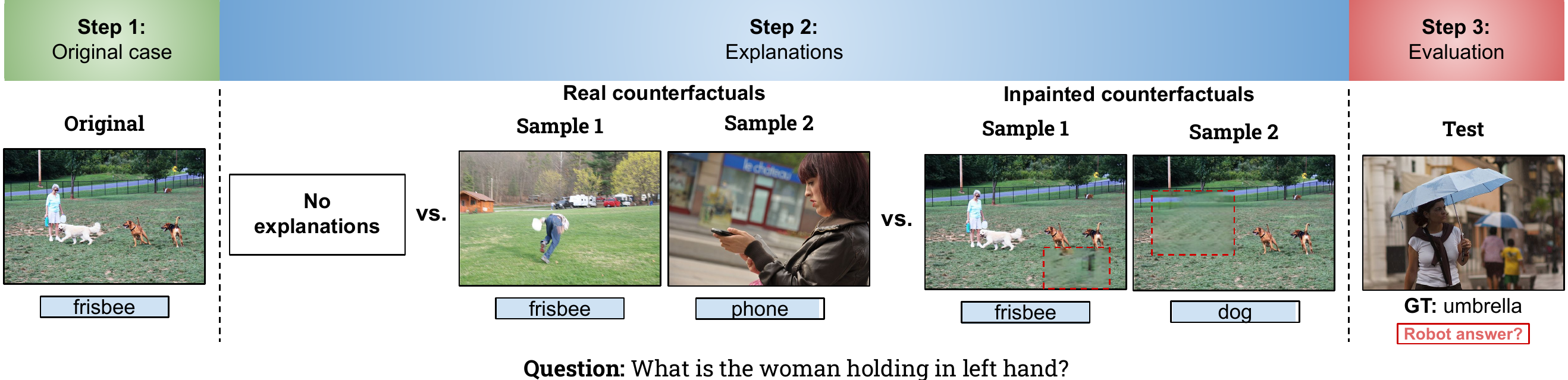}}
\caption{The workflow for different groups of the study. While steps 1 and 3 are shared among groups, the explanation step differentiates between them. In in-painted counterfactuals, samples 1 and 2 are in-painted over the least attended and most attended areas respectively. The real counterfactual images are sampled from the VQA dataset.}
\label{fig:exp_correctness_prediction_interfaces}
\end{figure*}

Note that in all cases, we make sure all images are relevant to the question asked since VQA models are not trained to answer irrelevant questions about images \cite{ray2016question}. 
Figure \ref{fig:exp_correctness_prediction_interfaces} visualizes the different interfaces used for the CG-NoExp group and the counterfactual groups.

%Moreover, we also would like to examine the in-painted counterfactuals by comparing them to a different form of explanation. In that regard, we define two groups with non-in-painted samples: CF-AltImg and CG-RandIQ. In the CF-AltImg group, we keep the same question in the samples the original, but the image is changed to another real image. Since the alternative images are fetched from IQ pairs in the VQA dataset, they still have a context relevant to the question. We do not test the case where an irrelevant image is asked to the question since VQA models are not trained to answer irrelevant questions well . In the CG-RandIQ group, we provide the samples from random IQ pairs in the dataset regardless of the original IQ. These two groups improve our view of a baseline so we can actually examine the role of in-paintings on user's mental model. 

\begin{center}
\begin{table*}[t]%
\caption{User study groups for answer correctness prediction.\label{table:accuracyprediction_groups}}
\centering
\begin{tabular*}{500pt}{@{\extracolsep\fill}lccD{.}{.}{3}c@{\extracolsep\fill}}
\toprule
\textbf{Group}
&\multicolumn{2}{@{}c@{}}{\textbf{Samples}} & \textbf{Attention source} \\\cmidrule{2-3}
 & \textbf{Images} & \textbf{Questions} \\
\midrule
CG-NoExp & \text{---} & \text{---}  & \text{---} \\
CG-RandIQ & Random images & Random questions  & \text{---}  \\
CF-HAT & Original image in-painted over least/most attended areas & Original question  & \text{Human annotations}  \\
CF-AIAtt & Original image in-painted over least/most attended areas & Original question  & \text{AI attention}  \\
CF-AltImg & Alternative real images & Original question  & \text{---} \\
\bottomrule
\end{tabular*}
\end{table*}
\end{center}

%In all groups, users start by viewing AI's answer and confidence to the original IQ. They also end each episode in the evaluation section where they attempt to predict AI's correctness when answering the same question for a different image. This evaluation provides a metric of how well the users can predict AI behaviour based on one sample. When comparing the CG-NoExp to other groups we examine the general impact of counterfactuals. On the other hand, by comparing in-painting groups with real counterfactuals, we can assess the role of in-painting in generating counterfactuals (figure \ref{fig:exp_correctness_prediction_interfaces}). As mentioned earlier, the in-paintings are over the least attended and most attended areas of the image. By comparing the user performances between groups CF-HAT and CF-AIAtt we can investigate the role of attention source in producing helpful counterfactuals.

%\begin{figure*}[h]
%\centerline{
%\includegraphics[width=500pt]{images/no_exp_vs_inpaint_grp_sample.pdf}}
%\caption{The interfaces for the experiments that evaluates the impact of in-painted counterfactuals in the task of answer correctness prediction.}
%\label{fig:exp_config_noexp_vs_in-paint}
%\end{figure*}
 
%\begin{figure*}[h]
%\centerline{
%\includegraphics[width=500pt]{images/real_vs_inpaint_grp_sample.pdf}}
%\caption{The interfaces for the experiments that compare the impact of in-painted counterfactuals vs. the real image counterfactuals.}
%\label{fig:exp_config_real_vs_in-paint}
%\end{figure*}

%In this experiment, we evaluated the role of attention sources in generating effective counterfactuals.\\
%In our experiment, we compared the performance of the users based on the source of attention that selects the counterfactual in-painting regions. in the HAT group, the source is the human attention, and in the AI group, the source is AI object attention.\\

\vspace{-5mm}
\section{Results and discussion}

In this section, we cover the results from the user studies conducted for the two major tasks described previously: answer change prediction and answer correctness prediction. 
%For the answer change prediction, we compare the user accuracy with and without the counterfactual samples. 
%The study is conducted in three subgroups based on the areas of in-painting in the sample and test counterfactual images. While the images in these experiments are altered, the question for each episode stays the same.

\subsection{Answer Change Prediction}
Table \ref{table:anschange_results} provides detailed numbers on the user accuracy in all groups. For each group, the users collectively predict a certain number of episodes which is outlined as N in the table. 
We show the accuracy of users correctly predicting the system would be INCORRECT for the cases when the VQA model is INCORRECT and similarly for when the VQA model is CORRECT. In the last column, we finally present the normalized accuracy which is the average of the accuracy for the CORRECT and INCORRECT cases. 
Since the number of correct cases is more than the number of incorrect cases for a VQA model, a normalized accuracy score mitigates potential spurious increases of accuracy simply because a user always predicted a model would be correct. If a user always predicted a model would be correct, the recall for CORRECT cases would be 100\% and 0\% for the INCORRECT cases, resulting in a normalized accuracy of only 50\%.    
%Aside from the overall accuracy in each subgroup which is presented as User Acc in the table, we also compared the normal accuracy in each group. The normal accuracy is the average between accuracy rates when the AI answer changes and when it does not. By providing this normalized metric we can mitigate the bias imposed by the ratio of answer change in the test in-paintings.\\
\begin{center}
\begin{table*}[h]%
\caption{Normalized user accuracies in answer change prediction task.\label{table:anschange_results}}
\centering
\begin{tabular*}{300pt}{@{\extracolsep\fill}lll}
\toprule
\textbf{Group} & \textbf{Baseline} & \textbf{Counterfactual} \\
\midrule
CF-MinAtt & 54.23\% & $\textbf{62.52\%} ^ { *}$ \\
CF-MidAtt & 56.29\% & $\textbf{66.38\%} ^ { **}$\\
CF-MaxAtt & 62.91\% & $\textbf{66.01\%} ^ { **}$\\
All & 61.30\% & $\textbf{66.98\%} ^ { ***}$ \\
\bottomrule
\end{tabular*}
\end{table*}
\end{center}

\xhdr{Counterfactual examples help over not providing examples for predicting answer change}
Users exposed to the counterfactual samples can predict the answer change better than the users in the baseline group. Moreover, we observe a consistent improvement in all subgroups regardless of the in-painting patterns. Even in CF-MinAtt and CF-MidAtt users tend to do better when exposed to the counterfactuals although predicting an answer change for those cases can be inherently harder. These findings suggest a positive impact by the counterfactual samples on the mental model independent of the testing scenario.

%The top finding in the answer change prediction task is the consistency of improvements across all subgroups.  
%This improvement is present in the normalized accuracy as well as the overall accuracy which indicates a better mental model regardless of the probability of answer change for the users. \\

\subsection{Answer Correctness Prediction}

Here, we evaluate the user's accuracy in predicting whether a VQA model will be correct or not on an unseen test image. 
We conduct most of our experiments on this task since this task is more realistic and challenging. Our ultimate goal is to see if users can learn from counterfactual explanations to predict the model's performance on unseen images. 
%This task can naturally be more challenging since the users are asked about AI's answer on an image completely different than the original image. Providing accurate predictions for AI's performance on arbitrary inputs can provide a strong metric for the user's mental model of the system.\\
We divide the correctness prediction results into two subgroups based on cases where the VQA model is CORRECT and INCORRECT as shown in Table \ref{table:correctness_results}. 
Users tend to have an initial optimistic bias towards AI accuracy and as a result, they are more inclined to predict that the AI machine would be correct. As described previously, to prevent a spurious accuracy increase simply due to a user predicting a model will be correct more often, we compute the normalized user accuracy as an average between AI correct and AI incorrect cases.

\begin{center}
\begin{table*}[h]%
\caption{User accuracy in answer correctness prediction task.\label{table:correctness_results}}
\centering
\begin{tabular*}{300pt}{@{\extracolsep\fill}lcccccccD{.}{.}{3}c@{\extracolsep\fill}}
\toprule
\textbf{Group} & \multicolumn{2}{@{}c@{}}{\textbf{AI correct}} & \multicolumn{2}{@{}c@{}}{\textbf{AI incorrect}} \\ \cmidrule{2-3}\cmidrule{4-5}
& N & User Acc. & N & User Acc. & Norm. Acc.\\
\midrule
a. CG-NoExp & 2995 & 70.42\% & 2605 & 45.60\% & 58.01\%\\
b. CG-RandIQ & 2921 & 78.26\% & 2659 & 46.30\% & 62.28\%\\
c. CF-HAT & 3005 & 73.14\% & 2625 & \textbf{54.48\%} & 63.81\%\\
d. CF-AIAtt & 2942 & 77.16\% & 2558 & 42.81\% & 59.98\%\\
e. CF-AltImg & 1643 & \textbf{85.88\%} & 917 & 43.84\% & \textbf{64.86\%}\\
\bottomrule
\end{tabular*}
\end{table*}
\end{center}

%\begin{figure*}[h]
%\centerline{
%\includegraphics[width=250pt]{images/acc_pred_chart.pdf}}
%\caption{}
%\label{fig:exp_correctness_prediction_chart}
%\end{figure*}

\noindent We now summarize our findings:

\xhdr{Counterfactual examples help over showing no examples}
All counterfactual groups - CF-AltImg, CF-HAT, CF-AIAtt - show improvement over the control group where no explanations are shown for users' mental model as shown in Table \ref{table:correctness_results} row a vs. rows c,d,e. This is hardly a surprising result since counterfactual examples provide more information.

\xhdr{Counterfactual examples help over showing random examples}
To check how much we gain in the mental model from simply providing more information, we check the performance of users when we show two random examples to the users. We see that the counterfactual groups CF-AltImg and CF-HAT both improve the mental model over simply showing random examples as shown in Table \ref{table:correctness_results} row b vs. rows c and e. This shows that counterfactuals are indeed an effective form of showing examples of how a model behaves to users. 

\xhdr{Generated counterfactual images can be a close substitute to realistic counterfactual images}
We see that a generated counterfactual image using an in-painting network \cite{chang2018explaining} based on human-annotated attention (row c of Table \ref{table:correctness_results}) can be almost as effective as a real retrieved counterfactual image from the VQA dataset (row e). While human-attention annotation is still currently needed, it is a step towards automating the counterfactual generation process.

\xhdr{Fully automating the generation process for counterfactual images can be tricky and currently doesn't seem to help mental model improvement}
As seen from row d of Table \ref{table:correctness_results}, if we use the model's attention values to decide objects to remove, the counterfactual images generated do not improve the user's mental model significantly over no example cases or when random cases are shown. This suggests that further research is needed to effectively automate the counterfactual generation process.\\ 
Overall, the results indicate that counterfactual explanations have a positive impact on the user mental model. While studies on case-based explanations\cite{kenny2021explaining,keane2020good} have shown random examples can improve users' mental models, our results indicate that controlled counterfactuals can better improve the mental model with the same number of examples shown. In certain application fields such as medicine, data is expensive, and hence, counterfactuals can help achieve an increase in mental models with fewer data points than showing random examples. We also see that GAN-generated counterfactual examples show comparable efficacy when evaluated against real retrieved counterfactual examples. However, our best-performing GAN-generated counterfactual relies on human-annotated attention maps being available. Further research needs to be conducted to explore effective ways of generating GAN counterfactuals without the need for human attention for the GAN-based method to be scalable. 

%\xhdr{}
%On the other hand, the in-paintings based on human attention (CF-HAT) make the users more skeptical of AI performance and help them to score better when AI is inaccurate. The comparison between the AI-based counterfactuals (CF-AIAtt) and human-based counterfactuals (CF-HAT) suggests the human attention as a more helpful source to produce effective counterfactuals.

%Overall, our findings show the CF-AltImg as the most successful group in predicting AI correctness. The results suggest that in this group users tend to trust the system better. Essentially, this observation shows that the users build a higher trust in AI's success when they see its performance across different images. In other words, the users trust the system to succeed for a different image if they observe the system in samples with different images rather than the same image with in-paintings. This impact on users' mental model becomes more obvious as we compare these groups with CF-RandIQ, where both the image and the question are altered. Although the level of trust in this group is still higher than the in-painting groups, it does not reach the level of CF-AltImg.\\

\section{Conclusion}

%In this work, we investigated the approaches to produce counterfactual explanations for a VQA system. To quantify the impact of explanations on user mental model we conducted a series of studies that evaluate the counterfactual explanations helpfulness based on users ability in predicting AI agents behavior. Based on this analogy we defined two prediction tasks and examined the effect of different counterfactual example generation methods in assisting user's prediction.\\
%Our results indicate a consistent improvement of user's accuracy when predicting AI's answer change in a counterfactual test. This improvement is independent of the attention weight of the in-painted area and demonstrates the positive impact of counterfactual in-paintings on user mental model.\\
%In the correctness prediction study, we present a more challenging task to examine the impact of counterfactual explanations on predicting AI's correctness. Our experiments compare the in-painting counterfactuals with alternative explanations. The results from this study shows that real image counterfactuals tend to bias the users to over-trust the VQA system and reduce their prediction accuracy in cases where the AI is incorrect. On the other hand, the in-painting counterfactuals helped users the most in predicting AI incorrect responses. Moreover, a comparison between the AI attention and human attention shows that unlike the common approaches that use AI attention to produce explanations, human attention provides a better source in producing counterfactual explanations.
% Future directions?
In this work, we demonstrated that showing counterfactual images is helpful for the mental model improvement of users in predicting a VQA model's performance. We showed that counterfactual examples are more effective than showing random examples or not showing any examples at all. We also showed that a generative approach to generate counterfactual images can also be effective at improving the mental model of users. Investigating different image editing methods and also the impact of the counterfactual quality on user mental mind can serve as interesting topics for the next steps of this work. We hope these results can serve as a foundation to improve generative models for producing effective counterfactual explanations to improve user mental models for the safe and effective deployment of AI systems in the wild.

%\backmatter
\section*{Acknowledgments}
This research was developed with funding from the Defense Advanced Research Projects Agency (DARPA) under the Explainable AI (XAI) program. The views, opinions and/or findings expressed are those of the author and should not be interpreted as representing the official views or policies of the Department of Defense or the U.S. Government.

%\subsection*{Financial disclosure}
%None reported.

\subsection*{Conflict of interest}
The authors declare no potential conflict of interests.

\nocite{*}% Show all bib entries - both cited and uncited; comment this line to view only cited bib entries;
\bibliography{main}%

\clearpage

\section*{Author Biography}

\begin{biography}{\includegraphics[width=76pt,height=86pt]{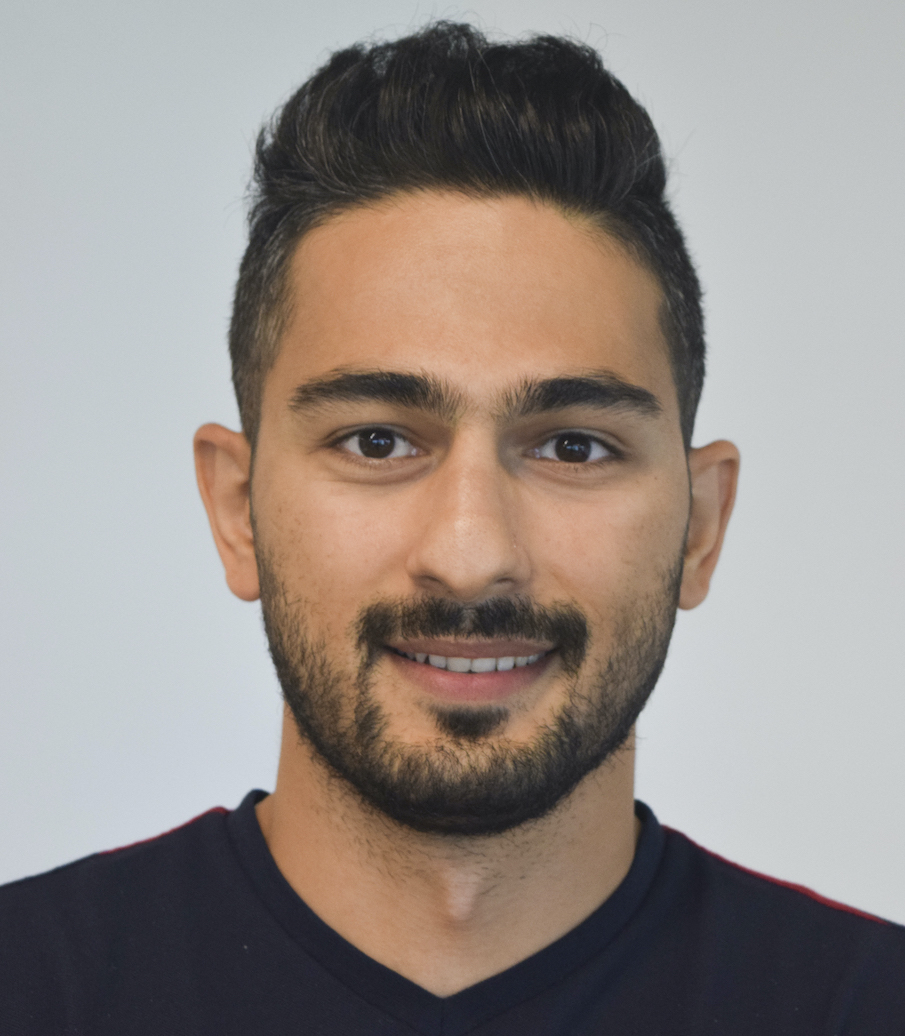}}{\textbf{Kamran Alipour} is a Ph.D. candidate in the Computer Science Department of the University of California, San Diego. He has a B.Sc. degree in Aerospace Engineering (2011) and an M.Sc. degree in Aerospace Engineering (2013). His research involves explainable AI and human-computer interaction. He specifically works on causal explanations and their helpfulness in human-AI collaboration tasks.}

\end{biography}

\end{document}